\documentclass[letterpaper, 10 pt, conference]{ieeeconf}

\IEEEoverridecommandlockouts
\usepackage{adjustbox}
\usepackage{ulem}
\usepackage{bm}

\usepackage[english]{babel}

\usepackage{multicol}
\usepackage{booktabs}
\usepackage{tabularx}
\usepackage[font=normalsize,labelfont=bf]{caption}
\usepackage [autostyle, english = american]{csquotes}
\MakeOuterQuote{"}
\usepackage[protrusion=false,stretch=10,shrink=10]{microtype}
\usepackage{amsmath, amssymb, amsfonts}
\usepackage{graphicx}
\usepackage{dblfloatfix}   
\usepackage{caption}
\usepackage{subcaption}
\usepackage{siunitx}
\let\labelindent\relax
\usepackage{enumitem}
\usepackage{listings}
\usepackage{booktabs}
\usepackage{textcomp}

\usepackage{textcomp}

\newcommand{\redwrap}[1]{{#1}}

\usepackage{xcolor}
\definecolor{myindigo}{HTML}{332288}
\definecolor{mycyan}{HTML}{88CCEE}
\definecolor{myteal}{HTML}{44AA99}
\definecolor{mygreen}{HTML}{117733}
\definecolor{myolive}{HTML}{999933}
\definecolor{mysand}{HTML}{DDCC77}
\definecolor{myrose}{HTML}{CC6677}
\definecolor{mywine}{HTML}{882255}
\definecolor{mypurple}{HTML}{AA4499}
\definecolor{mygray}{HTML}{DDDDDD}

\title{
\LARGE \bf
Oh F**k! How Do People Feel about Robots that Leverage Profanity?
}

\author{Madison R. Shippy$^1$, Brian J. Zhang$^1$, Naomi T. Fitter$^1$%
\thanks{$^{1}$Collaborative Robotics and Intelligent Systems (CoRIS) Institute, Oregon State University, Corvallis, Oregon, USA. {\tt \small fittern@oregonstate.edu}}
}

\begin{document}

\maketitle

\thispagestyle{empty}
\pagestyle{empty}

\begin{abstract}
Profanity is nearly as old as language itself, and cursing has become particularly ubiquitous within the last century. At the same time, robots in personal and service applications are often overly polite, even though past work demonstrates the potential benefits of robot norm-breaking. Thus, we became curious about robots using curse words in error scenarios as a means for improving social perceptions by human users. We investigated this idea using three phases of exploratory work: an online video-based study ($N = 76$) with a student pool, an online video-based study ($N = 98$) in the general U.S. population, and an in-person proof-of-concept deployment ($N = 52$) in a campus space, each of which included the following conditions: no-speech, non-expletive error response, and expletive error response. A surprising result in the outcomes for all three studies was that although verbal acknowledgment of an error was typically beneficial (as expected based on prior work), few significant differences appeared between the non-expletive and expletive error acknowledgment conditions (counter to our expectations). 
Within the cultural context of our work, the U.S., it seems that many users would likely not mind if robots curse, and may even find it relatable and humorous. This work signals a promising and mischievous design space that challenges typical robot character design.
\end{abstract}


       




\vspace{-0.05in}
\section{Introduction}
\vspace{-0.05in}

\begin{figure}[t]
    \centering
    \vspace{-0.37in}
    \includegraphics[width=0.8\columnwidth]{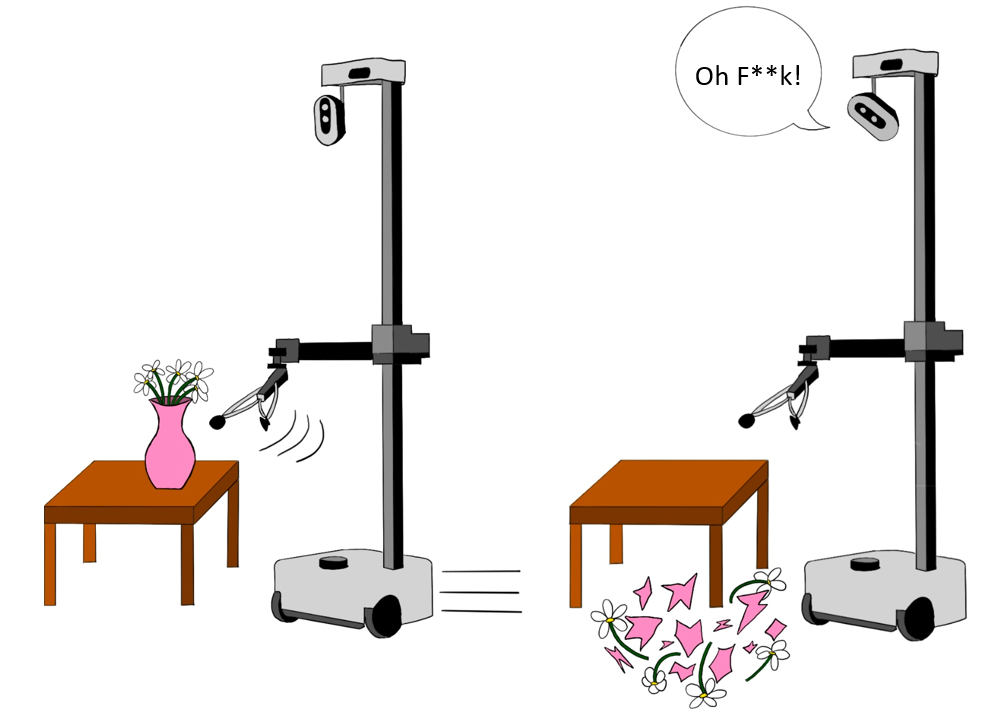}
    \vspace{-0.05in}
     \caption{Depiction of the Hello Robot Stretch mobile manipulator responding to an accident with an expletive.
     }
     \label{fig:strech_splash}
     \vspace{-0.2in}
 \end{figure}

As robots enter increasingly human-populated settings, a broader exploration of robot emotional expression, and even mischief, may help these systems to succeed, especially in challenging cases such as interaction repair after errors. Although many robots in personal and service roles lack apparent emotion, and in particular avoid expressing negative emotion, robots that do show emotion seem more anthropomorphic~\cite{spatola2021ascribing} and are more likely to be forgiven when they fail at a task~\cite{yam2021robots}. 
Despite these potential benefits, there is little past research on using uncouth robot behavior (such as cursing in response to a mistake) to support improved interactions with interlocutors. Accordingly, in this exploratory work, we aimed to understand gains that might come from robots using profane exclamations when failing at a task in a mock service setting (e.g., Fig.~\ref{fig:strech_splash}).

Although the idea of a cursing robot improving user perception may at first seem counterintuitive, further past work highlights how such a behavior can benefit interactions with people. 
Beyond the examples above, prior research showed that robots with apparent emotion were perceived as more trustworthy~\cite{bartneck2003interacting} and as better teammates~\cite{tapus2007socially}. Robot violation of social norms also gives the illusion of agency and experience~\cite{yasuda2020perceived}. Of course, certain social norm-breaking by a robot may be considered offensive, or even create long-lasting negative perceptions of a robot; however, at least some past work supports the idea that cursing is a social norm violation that does not create a long-lasting emotional response, and that it is seen as more humorous than offensive~\cite{yasuda2020perceived} \redwrap{(at least within the U.S., and potentially in selected other cultural contexts)}. Thus, we chose to study cursing robots as a potential strategy for simultaneously harnessing the benefits of emotional and norm-breaking robots without unduly offending users. Based on the related work, we saw potential in cursing in response to error.

The presented exploratory work details three studies of a cursing robot: an online video-based study with student participants, an online video-based study on a general adult population, and an in-person proof-of-concept deployment on a college campus. 
Our central research goal was to \emph{investigate whether and how the under-studied idea of profanity in robot speech can lead to benefits in interaction with human users}. 
After a digest of related work in Section~\ref{Related Work}, Sections~\ref{initial}, Section~\ref{follow up}, and Section~\ref{in_person} describe the methods and results of the first online study (with U.S. university students), follow-up video-based study (with a broader U.S.-based sample), and exploratory in-person deployment, respectively. We discuss strength, limitations, and take-aways from the work in Section~\ref{Discussion}. 
The primary contributions of this work include a replicated study that supports the idea of using robot reactions to a mistake (including both non-profane and profane utterances) to positively impact interlocutor perceptions, as well as grounding for further research in the area of uncouth robots.


\vspace{-0.05in}
\section{Related Work}
\label{Related Work}
\vspace{-0.05in}

To better contextualize our work, this section presents prior research from the fields of psychology and human-robot interaction (HRI) concerning why humans curse, what social benefits and pitfalls cursing can cause, and what happens when robots show emotion (or even violate norms).

\vspace{-0.05in}
\subsection{Cursing in Humans}
\vspace{-0.05in}
To understand prospective social robot cursing (or swearing), we must first understand some of the underlying intricacies of human communication. 
One past paper defines swearing as expression that: a) refers to something that may be stigmatized in the speaker's culture, b) should not be taken literally, and c) can be used to strongly express emotion~\cite{monaghan2012cultural}. Humans have been cursing since the dawn of developed language~\cite{vingerhoets2013swearing}. Cursing has been on the rise since the 1960's, with its power and effect gradually decreasing as it becomes more commonplace in language~\cite{baruch2007swearing}. The use of this type of profane language falls on a spectrum from unconscious/automatic to conscious/controlled~\cite{jay2009utility}. When humans swear on the unconscious side of the spectrum, there is a catharsis effect meant to inhibit outward aggression~\cite{pinker2007stuff} or satisfy a primal instinct (similar to growling in animals)~\cite{patrick1901psychology}. One study found that anger, frustration, and humor were the most commonly reported emotions around cursing~\cite{jay2006memories}. 

Humorous cursing can closely inform our work. One study posits that humor is commonly a form of social risk-taking, and found that curse words were positively correlated with aggressive humor styles and affiliative humor styles~\cite{kennison2019humor}. Other work confirms that swearing is often used for humorous effect and promotes feelings of group inclusion~\cite{vingerhoets2013swearing, baruch2007swearing}.

Cursing can also be an important part of human social bonds. A study in a New Zealand soap factory examined the conversations between colleagues and found a high level of cursing within the team~\cite{daly2004expletives}. Collective swearing, when related to job frustration, led to feelings of social connectedness in the group. This same pattern is seen in adolescents as a sign of solidarity in social groups~\cite{stapleton2010swearing}. These findings suggest that cursing can help express positive group identity, although both studies acknowledged that swearing can also have (context-dependent) negative consequences. 

This prior work shows that cursing can yield physiological, humorous, and social benefits in humans. However, the social benefits can be context- and content-dependent. We aimed \emph{to assess if the social benefits (and pitfalls) of cursing also apply in interactions with robots.}

\vspace{-0.05in}
\subsection{Emotional Response and Norm Violation}
\vspace{-0.05in}
The ideas of apparent emotions displayed by robots and responses to robot norm-breaking are important to our work. When roboticists create the impression of emotions in robotic systems, people tend to be more forgiving of the robot's mistakes~\cite{yam2021robots}; this is especially true for physically embodied robots~\cite{kontogiorgos2020embodiment}. For example, one study found that apparent emotions by a service robot led to higher customer satisfaction after a mistake~\cite{yam2021robots}. Making a mistake and acknowledging the error is a valuable learned skill in social creatures, as past human psychology work shows that acknowledgement of mistakes is predictive of others' forgiveness of mistakes~\cite{schutte2001emotional}. In the robotics realm, one project likewise found that robots were more easily forgiven when using affirming humor to acknowledge a mistake~\cite{green2022s}. We considered robot cursing after an error to be a potential tactic for acknowledging error and expressing apparent emotion. 

When creating a robot that curses, it is also important to consider the potential benefits of social norm violation. In the current state of the art, robots are usually polite and avoidant of anger or frustration that might violate norms. However, in human-human interactions, norm-breaking enhances feelings of group inclusion if the norm is not perceived to be critical~\cite{popa2014positive}. A good example of a non-critical norm violation is cursing, which has become ubiquitous in many cultural contexts (e.g., the U.S.)~\cite{baruch2007swearing}. While there is limited work on robots that break social norms, past examples show that robots that use disparaging or crass humor are better than humans at getting away with it~\cite{tay2016types}, and that lightly roguish robotic comedians can succeed in live performance~\cite{vilk2020comedians}. 
Another study considered social norm violation using three conditions in the context of a card game: insulting, cheating, and cursing~\cite{yasuda2020perceived}. While participants in the cheating and insulting category had negative feelings that lasted after the norm-breaking, the ones in the cursing condition 
found the cursing amusing, which reinforced our selection of cursing in particular as the robot norm-breaking to probe in our study.

The prior work shows that apparent emotion, acknowledgment of error, and non-critical social norm violation can enhance perceptions of personal and service robots, which have many opportunities to fail gracefully in regular operations.
We aimed to \emph{study cursing in particular as a method for acknowledging error and (perhaps even somewhat counterintuitively) enhancing user perceptions of robots.}



\vspace{-0.05in}
\section{Initial Online Study (University Students)}
\label{initial}
\vspace{-0.05in}

In this exploratory study, we aimed to assess participant opinions of a robot that display apparent emotion and break social norms by cursing upon making a mistake. 
We used an online video-based study, a reasonable method for beginning to compare different variants of robot behaviors~\cite{gosling2004should}, in a \redwrap{U.S.-based} university student pool.

\vspace{-0.05in}
\subsection{Methods}
\label{Methods 1}
\vspace{-0.05in}
The methods of the within-subjects video-based study described below were approved by the Oregon State University institutional review board. 

\subsubsection{Study Design}
We designed the study around the Hello Robot Stretch RE2, a modern mobile manipulator and representative robot for personal/service robotic applications.

The study included three conditions:
\begin{itemize}
    \item \emph{No-Speech:} In this control condition, the robot did not react verbally to errors.
    \item \emph{Non-Expletive:} In this experimental condition, the robot verbally reacted to errors with a non-expletive.
    \item \emph{Expletive:} In this experimental condition, the robot verbally reacted to errors with an expletive exclamation.
\end{itemize}

Responses to the proposed type of robot exclamation may be context-dependent. Accordingly, we presented three types of robot failure (with great relevance to personal and service robots) in the study, as described below:
\begin{itemize}
    \item \emph{Bumping:} Trying to pass a table, but bumping it.
    \item \emph{Knocking:} Trying to grasp a cup, but knocking it over.
    \item \emph{Dropping:} Trying to grasp a pen, but dropping it.
\end{itemize}
The failure in each video gave the robot a visible mistake to respond to. We treated these three error presentations as repetitions of each study condition.



\subsubsection{Stimulus Creation}
\label{Stimuli Design}
\begin{figure}[t]
    \centering
    \vspace{0.05in}
    \includegraphics[width=0.9\columnwidth]{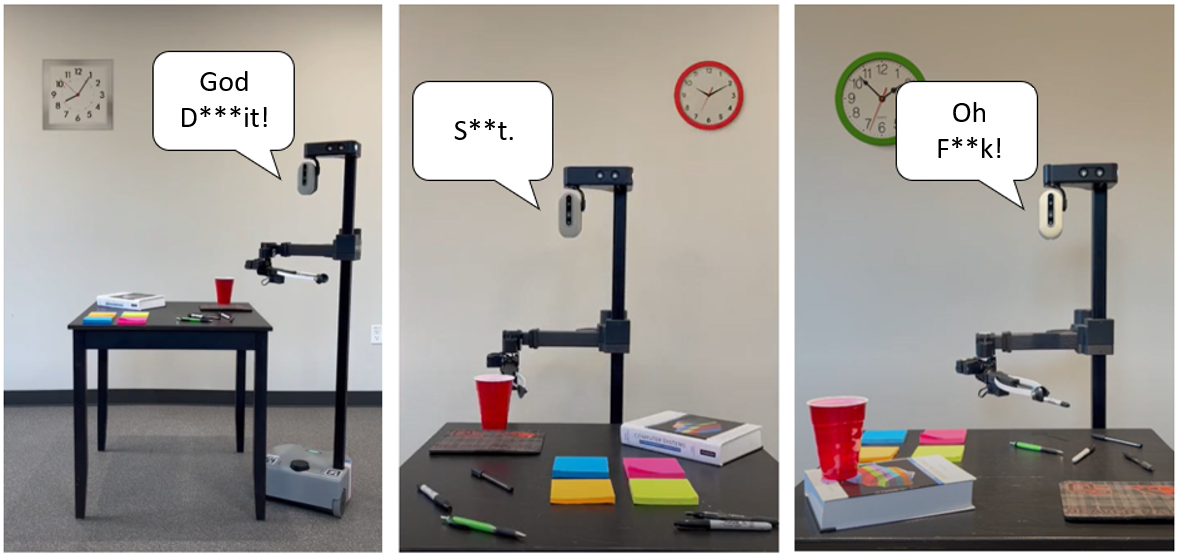}
    \vspace{-0.05in}
     \caption{Still frames from video stimuli for bumping the table (left), knocking over the cup (center), and dropping the pen (right). The speech bubbles were added for this figure.
     }
     \label{fig:vid_stills}
     \vspace{-0.05in}
 \end{figure}

To generate the video stimuli, we recorded video of the three staged robot errors mentioned above. The staging for these error scenarios was a well-lit, otherwise empty room with the Stretch robot, a table, books, pens, sticky notes, and a cup in frame. We directly teleoperated the robot while filming three videos in total, one of which represented each type of error. 
In all three videos, the ``mistake'' was followed by the robot moving its camera toward the affected object to help draw attention to the mistake that has just occurred (a key context for the ensuing declarative). 
Individual frames from these base three videos appear in Fig. \ref{fig:vid_stills}. All videos were roughly 9s long. 

To avoid potential confounds arising from differences in robot motion, error presentation, etc., we video-edited each of the three base videos to generate nine total stimuli, one for each cross of condition and error scenario.
We selected six phrases, three with non-expletive exclamations and three with expletive exclamations, for the video creation, as shown in Table~\ref{tab:phrase}. The phrases were rendered as speech using the Amazon Polly ``Brian'' voice. We selected a male-presenting voice since past studies have shown that it is more socially acceptable for men to curse (compared to women)~\cite{edelsky1976subjective,popp2003gender}. 
We audio-edited the expressions into the videos atop the natural sound of the robot and environment, leaving the three control condition videos with no speech. 
We mitigated the chance that the participants would notice they were watching the same base videos multiple times by adding different colors and shapes of clocks to the background. The final video stimuli are available on GitHub~\cite{github}. 

\begin{table}[t]
    \centering
    \caption{Table showing the three expletive and three non-expletive phrases used in the stimuli for each behavior.}
    \vspace{-0.05in}
    \label{tab:phrase}
    \begin{tabular}{c|c c}
    \toprule
         Behavior & Non-Expletive & Expletive\\
         \midrule
           Bumping & ``Oh no!'' & ``Godd***it!''\\
           Knocking & ``Oops.'' & ``S**t.''\\
           Dropping & ``Ah drat!'' & ``Oh f**k!'' \\           
         \bottomrule
    \end{tabular}
    \vspace{-0.15in}
\end{table}


\subsubsection{Measures}
\label{Measures}
We sought to measure constructs such as perceived social warmth of the robot and feelings of interpersonal closeness between the robot and human onlooker using opening, post-stimulus, and demographic questionnaires. 

The \emph{opening questionnaire} included the Negative Attitudes Towards Robots Scale (NARS), which captures general attitudes toward \emph{interactions with robots}, \emph{social influence of robots}, and \emph{emotions in robots}~\cite{NARS}. We administered this question set using 6-pt. Likert-type scales from ``Strongly Disagree'' to ``Strongly Agree.'' Based on past examples in the literature~\cite{chyung2017evidence, weems2001impact}, we removed the ``Neutral'' category to elicit clearer participant opinions. 

The \emph{post-stimulus questionnaire} included four inventories: the \emph{Robotic Social Attributes Scale (RoSAS)}, the \emph{Inclusion of Other in Self (IOS) Scale}, the \emph{Joke Rating Scale (JRS)}, and a subset of the \emph{Godspeed Questionnaire}. The RoSAS assessed participant perceptions of \emph{warmth}, \emph{competence}, and \emph{discomfort}~\cite{RoSAS} using six component attributes on 6-pt. Likert-type scales from ``Definitely Not Associated'' to ``Definitely Associated.'' The IOS Scale captured how \emph{socially close} participants felt to the robot with seven standard images of increasingly overlapping Venn diagrams representing the ``self'' and ``other''~\cite{ios}. The JRS assessed robot \emph{humorousness} on eight subscales established in prior work~\cite{tay2016types}
using a 6-pt. Likert-type scale from ``Strongly Disagree'' to ``Strongly Agree.'' We also used the Godspeed \emph{anthropomorphism} and \emph{likability} scales
~\cite{godspeed, bartneck2023godspeed}; each has five semantic differential subscales, which we administered as 6-pt. scales.
    
The \emph{demographic questionnaire} recorded information about age, gender, nationality, ethnicity, experience with robots, STEM training, and comfort with profanity. 

\subsubsection{Procedure}
\label{procedure}
Prospective participants who opened our Qualtrics-based survey gave consent on the first page. The next page helped participants to set their speaker at the right volume level. After that, respondents completed the opening questionnaire. In the following part of the survey, participants began watching the stimuli. Participants watched one video at a time, each of which was followed by the post-stimulus questionnaire. The stimulus order was counterbalanced across participants in a constrained way such that alike behaviors or conditions would not play one after another. 
For each post-stimulus survey, participants were required to stay on the page for at least 30 seconds before the ``next'' button appeared. At the close of the survey, participants answered a free-response question about what stimuli aspects influenced their responses most 
and completed the demographic questionnaire. We administered an attention check question, which was answered correctly by all respondents. Respondents received course credit for participating in the 30-minute survey.

\subsubsection{Participants}
\label{init_particip}
Sufficient statistical power for this study was determined through an a priori G*Power analysis \redwrap{with an $f = 0.25$ effect size}; a sample size of approximately 60 was deemed suitable \redwrap{for} our study design. 
76 participants completed the survey (51 female, 21 male, 3 non-binary, and 1 queer masc). 
Participants were 18 to 41 years old ($\displaystyle M = 19.8, SD = 3.54$) and had a little experience with robots ($\displaystyle M = 2.2$ out of 5, $SD = 0.82$), but almost no experience with Stretch ($\displaystyle M$ = 1.1 out of 5, $SD = 0.46$). For the NARS scales (negative attitudes toward \emph{interactions with robots}, \emph{social influence of robots}, and \emph{emotions in robots}), participants were neutral with a slightly negative lean ($\displaystyle M = 3.2$ out of 6, $SD = 0.78$; $\displaystyle M = 4.0$, $SD = 0.83$; $\displaystyle M = 3.5$, $SD = 0.32$, respectively). 34\% of participants had STEM training. Participants were comfortable with profanity in their day-to-day lives ($\displaystyle M$ = 4.7 out of 6, $SD = 0.99$). 

\subsubsection{Analysis}
\label{analysis}
To assess differences in scale-wise questions, we used repeated measures analysis of variance (rANOVA) tests with an $\alpha=0.05$ significance level. In the case of significance, we conducted pairwise comparison tests with Tukey's HSD test. We report effect size using $\displaystyle \eta^{2}_{p}$. 


\vspace{-0.05in}
\subsection{Results}
\label{Results 1}
\vspace{-0.05in}
This section details the results (pre-planned and exploratory) of the post-stimulus questionnaire data. 

\subsubsection{Main Survey Results}

\begin{figure}[t]
    \centering
    \vspace{0.05in}
    \includegraphics[width=\columnwidth]{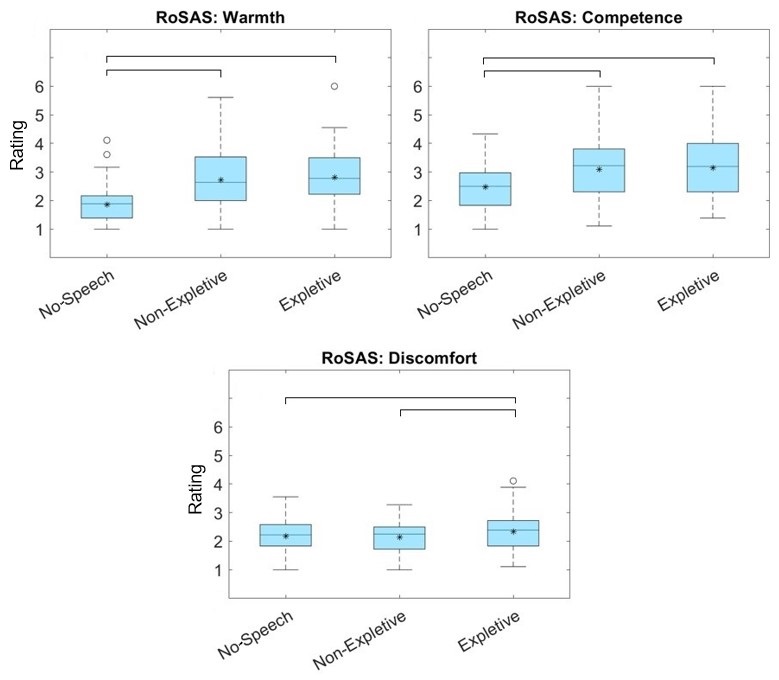}
    \vspace{-0.2in}
     \caption{Boxplots showing the survey responses for RoSAS warmth, competence, and discomfort in the initial study. In this figure and other boxplots throughout the paper, the boxes extend from the 25th to the 75th percentiles, the center horizontal line of the box marks the median, an asterisk (`*') marks the mean, whiskers show up to 1.5 times the interquartile range, and circles indicate outliers. Brackets indicate significant pairwise differences.
     }
     \label{fig:p_rosas}
     \vspace{-0.05in}
 \end{figure}

\emph{RoSAS} ratings 
appear in Fig.~\ref{fig:p_rosas}. We found significant main effects in the RoSAS \emph{warmth} ($F(2,150) = 65.9, p<0.001$, $\displaystyle \eta^{2}_p$ = 0.468), \emph{competence} ($F(2,150) = 42.7, p<0.001$, $\displaystyle \eta^{2}_p$ = 0.363), and \emph{discomfort} ($F(2,150) = 10.3, p<0.001$, $\displaystyle \eta^{2}_p$ = 0.121) rANOVA tests. Post hoc analysis of \emph{warmth} showed that the non-expletive and expletive conditions were rated significantly higher than the no-speech condition. \emph{Competence} likewise showed both speech conditions to appear more competent than the no-speech condition. For \emph{discomfort}, the expletive condition was rated higher than any other condition. \redwrap{We report post hoc test results succinctly here, but the full statistical reporting is available in the supplementary repository for the paper~\cite{github}.}


\begin{figure}[t]
    \centering
    \includegraphics[width=\columnwidth]{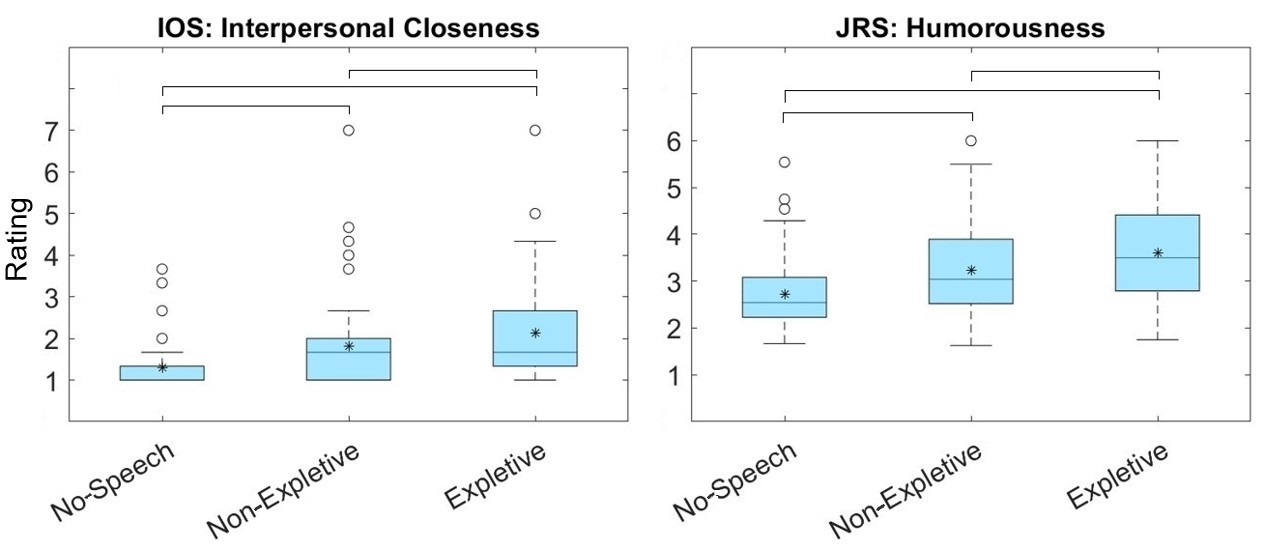}
    \vspace{-0.2in}
     \caption{Survey responses for IOS interpersonal closeness and JRS humorousness in the initial study.
     }
     \label{fig:p_ios_jrs}
     \vspace{-0.15in}
 \end{figure}

\emph{IOS} and \emph{JRS} ratings
are shown in Fig. \ref{fig:p_ios_jrs}. We found significant main effects in the \emph{interpersonal closeness} rANOVA test ($F(2,150) = 29.3, p<0.001$, $\displaystyle \eta^{2}_p$ = 0.281). 
There was a significant main effect for \emph{humorousness} ($F(2,150) = 47.8, p<0.001$, $\displaystyle \eta^{2}_p$ = 0.389). For both scales, all pairwise comparisons were significant; the expletive condition was rated most highly, then non-expletive, then no-speech.

\begin{figure}[t]
    \centering
    \vspace{0.05in}
    \includegraphics[width=\columnwidth]{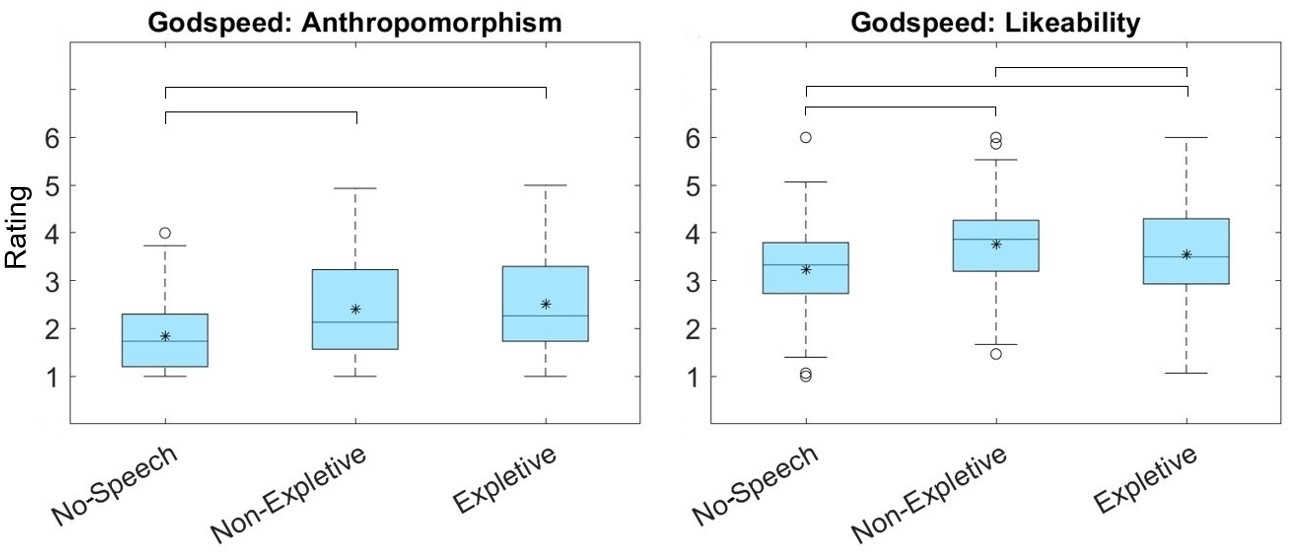}
    \vspace{-0.2in}
     \caption{Survey responses for Godspeed anthropomorphism and likeability in the initial survey.
     }
     \label{fig:p_godspeed}
     \vspace{-0.15in}
 \end{figure}

\emph{Godspeed} ratings 
appear in Fig.~\ref{fig:p_godspeed}. We found significant main effects for \emph{anthropomorphism} ($F(2,150) = 40.9, p<0.001$, $\displaystyle \eta^{2}_p$ = 0.353) and \emph{likeability} ($F(2,150) = 17.2, p<0.001$, $\displaystyle \eta^{2}_p$ = 0.187). Post hoc analysis for \emph{anthropomorphism} showed the non-expletive and expletive conditions to be rated higher than the no-speech condition. For \emph{likeability}, post hoc analysis revealed significance for all condition pairings; non-expletive was rated highest, then expletive, then no-speech. 

\subsubsection{Exploratory Analysis}

In the survey free-response data, we noticed that some participants were uncomfortable with the robot using a curse word with religious connotations (``Godd***it''). This led us to wonder if the multiple errors presented (i.e., bumping, knocking, and dropping) were functioning as true repetitions. We accordingly conducted an exploratory analysis in the form of 3$\times$3 2-way rANOVA (including both condition and scenario) to determine if there were significant differences in the perceptions of distinct error scenarios. We did not find significant differences between the error scenarios, but we did see an interaction effect between condition and scenario for \emph{discomfort} ($F(4,300) = 2.78, p = 0.027$, $\displaystyle \eta^{2}_p$ = 0.036) and \emph{likeability} ($F(4,300) = 2.51, p = 0.042$, $\displaystyle \eta^{2}_p$ = 0.032). Specifically, the video including ``Godd***it'' was perceived significantly more negatively than other stimuli on both of these scales.

We were curious if the removal of this particularly jarring curse word would lead to any differences in our results, so we removed the ``bumping'' scenario for all conditions from the dataset and re-ran the original one-way rANOVA test on this updated data.
Most of the main effect and pairwise testing results remained the same; we report on just the two scales with updated results here.
For \emph{discomfort}, we again found a significant main effect ($F(2,150) = 3.11, p = 0.048$, $\displaystyle \eta^{2}_p$ = 0.040). 
Compared to in the original results, however, the post hoc testing showed no pairwise significant effects for discomfort.
For \emph{likeability}, we again found a significant main effect ($F(2,150) = 21.8, p<0.001$, $\displaystyle \eta^{2}_p$ = 0.225). The pairwise comparison results, however, showed significant differences between both speech conditions and the no-speech condition, but no difference between non-expletive and expletive speech. 

\vspace{-0.05in}
\subsection{Summary of Key Findings}
\vspace{-0.05in}

The initial deployment most clearly showed benefits when the robot verbally acknowledged a mistake. Verbally responsive conditions were rated better on all scales other than \emph{discomfort}, where in the main results, expletive was the worst, and in the exploratory analysis, all conditions were evenly matched.
Participants generally did not care if that acknowledgment was a curse word, but some scales showed slight advantages and slight disadvantages of expletives. 
Specifically, the expletive condition yielded the highest \emph{interpersonal closeness} and \emph{humor} ratings, but for \emph{discomfort} and \emph{likeability}, the non-expletive condition was perceived best before the exploratory analysis. 
Within individual participants' input, we saw some free-response feedback that indicates that personalization may be important when designing foul-mouthed robots. For example, one respondent ``thought it was funny when [the robot] used profanity because it [made it] a little more lifelike,'' but another said ``if the robot said something with profane language then [it] seemed more aggressive and not as kind.''
At the same time, this type of robot behavior may be generally acceptable to younger people in the U.S. like those who responded to the survey; after the exploratory exclusion of the religiously-connotated curse, the expletive condition was better than or equal to the non-expletive condition on all fronts, 
suggesting that cursing robots merit further exploration.

\vspace{-0.05in}
\section{Follow-Up Prolific Study}
\label{follow up}
\vspace{-0.05in}

Compared to the general population, students may be more accepting of technology and profanity. Accordingly, in this follow-up study, we wanted to test if the online, video-based survey results replicated for a larger and more diverse sample. 

 \vspace{-0.05in}
\subsection{Methods}
\label{methods 2}
\vspace{-0.05in}




All aspects of the study methods were the same as in the previous study, except for the platform (Prolific.com), compensation (8.65 USD per hour), and participants.
98 \redwrap{U.S.-based} participants (51 female, 45 male, 1 transgender woman, and 1 transgender man) completed the survey. 
Participants were 18 to 41 years old ($\displaystyle M = 38.9, SD = 12.3$) and had little experience with robots ($\displaystyle M = 2.0$ out of 5, $SD = 0.73$), and almost no experience with Stretch specifically ($\displaystyle M$ = 1.1 out of 5, $SD = 0.39$). For the NARS scales, participants were neutral or had somewhat negative views of robots ($\displaystyle M = 2.8$ out of 6, $SD = 0.92$; $\displaystyle M = 3.7$, $SD = 1.08$; $\displaystyle M = 3.7$, $SD = 0.37$ on scales 1, 2, and 3, respectively). 41\% of participants had STEM training. Participants were comfortable with profanity in their day-to-day lives ($\displaystyle M$ = 4.3 out of 6, $SD = 1.22$). 


\vspace{-0.05in}
\subsection{Results}
\label{results 2}
\vspace{-0.05in}

This section details the results from the post-stimulus questionnaire data from the study. 

\subsubsection{Main Survey Results}

\begin{figure}[t]
    \centering
    \vspace{0.05in}
    \includegraphics[width=\columnwidth]{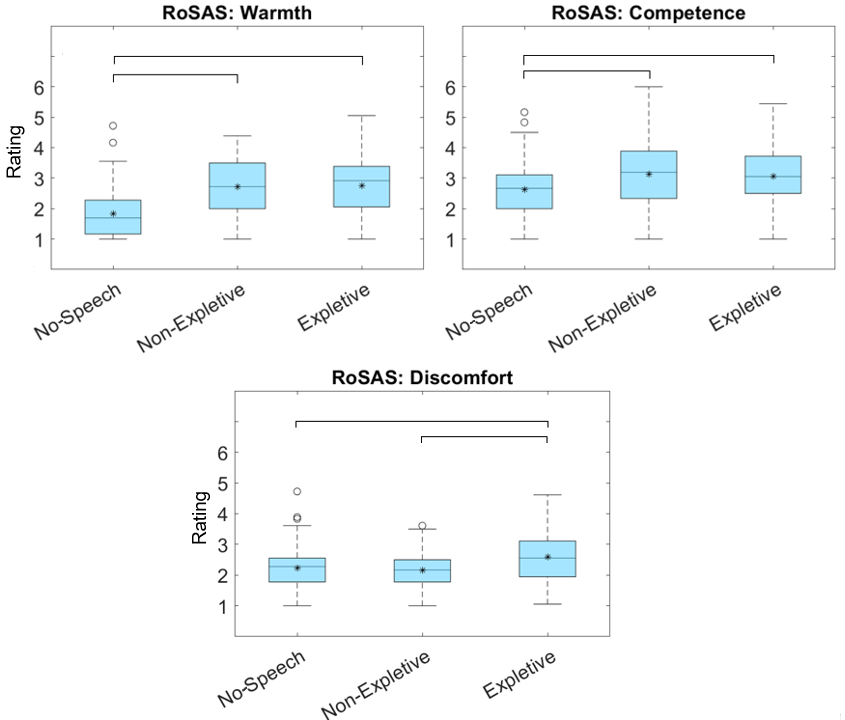}
    \vspace{-0.2in}
     \caption{Survey responses for RoSAS warmth, competence, and discomfort in the follow-up study.}
     \label{fig:f_rosas}
     \vspace{-0.1in}
 \end{figure}

Responses from the \emph{RoSAS} questionnaire 
are shown in Fig.~\ref{fig:f_rosas}. The RoSAS rANOVA results for the follow-up study yielded significant differences in \emph{warmth} ($F(2,194) = 138, p<0.001$, $\displaystyle \eta^{2}_p$ = 0.588), \emph{competence} ($F(2,194) = 48.2, p<0.001$, $\displaystyle \eta^{2}_p$ = 0.332), and \emph{discomfort} ($F(2,194) = 41.5, p<0.001$, $\displaystyle \eta^{2}_p$ = 0.300). Post hoc analysis of \emph{warmth} showed that the non-expletive and expletive conditions were rated significantly higher than the no-speech condition. \emph{Competence} ratings revealed the same patterns: both speech conditions appeared more competent than the no-speech condition. For \emph{discomfort}, expletive was rated significantly higher than any other condition.  


\begin{figure}[t]
    \centering
    \vspace{0.05in}
    \includegraphics[width=\columnwidth]{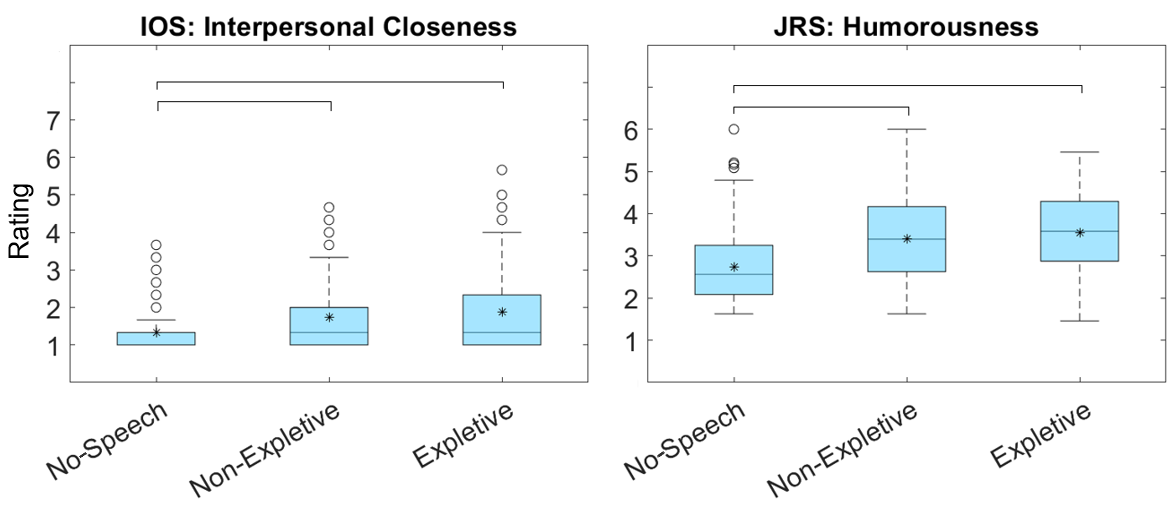}
    \vspace{-0.2in}
     \caption{Survey responses for IOS interpersonal closeness and JRS humorousness in the follow-up study.
     }
     \label{fig:f_ios_jrs}
     \vspace{-0.05in}
 \end{figure}

\emph{IOS} and \emph{JRS} ratings
appear in Fig. \ref{fig:f_ios_jrs}. The rANOVA test for the IOS scale showed that there was a significant effect between conditions ($F(2,194) = 25.5, p<0.001$, $\displaystyle \eta^{2}_p$ = 0.208). 
There was also a significant main effect for \emph{humor} ($F(2,194) = 47.8, p<0.001$, $\displaystyle \eta^{2}_p$ = 0.330). Post hoc analysis showed that the non-expletive and expletive conditions were rated significantly higher than the no-speech conditions on both scales.

  \begin{figure}[t]
    \centering
    \includegraphics[width=\columnwidth]{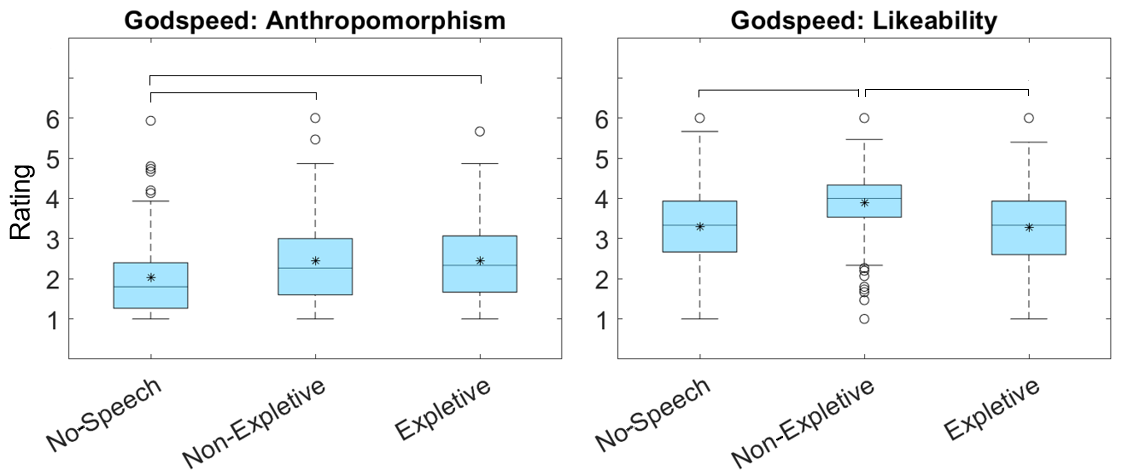}
    \vspace{-0.2in}
     \caption{Survey responses for Godspeed anthropomorphism and likeability in the follow-up survey.
     }
     \label{fig:f_godspeed}
     \vspace{-0.15in}
 \end{figure}


\emph{Godspeed} ratings
are shown in Fig. \ref{fig:f_godspeed}. We found significant differences for \emph{anthropomorphism} ($F(2,194) = 42.5, p<0.001$, $\displaystyle \eta^{2}_p$ = 0.305) and \emph{likeability} ($F(2,194) = 36.8, p<0.001$, $\displaystyle \eta^{2}_p$ = 0.275). The post hoc analysis for \emph{anthropomorphism} showed the expletive and non-expletive conditions to be rated higher than the no-speech condition. For \emph{likeability}, post hoc analysis revealed that the non-expletive condition appeared more likable than any other condition. 
\subsubsection{Exploratory Interaction Analysis}

In this follow-up study, we again noticed commentary in the free-response data about the stimulus using ``Godd***it,'' so we conducted the same 3$\times$3 2-way rANOVA (including both condition and scenario) exploratory analysis to identify differences in perceptions of the error scenarios. 
This time, there were significant interaction effects for \emph{discomfort} ($F(4,388) = 4.94, p<0.001$, $\displaystyle \eta^{2}_p$ = 0.048) and \emph{likeability} ($F(4,388) = 2.43, p = 0.047$, $\displaystyle \eta^{2}_p$ = 0.024); the stimulus including ``Godd***it'' was perceived most negatively. 
However, an exploratory one-way rANOVA test excluding the ``bumping'' scenario yielded no changes in significant effects (compared to the original analysis) for this follow-up study.

\vspace{-0.05in}
\subsection{Summary of Key Findings}
\vspace{-0.05in}
Almost all results replicated in the follow-up study, but results that did not replicate are important to the potential use of cursing robots in a general U.S. context.
Again, verbal responsiveness led to significant gains in all cases but \emph{discomfort}, for which expletive was worse than no-speech, and \emph{likability}, for which no-speech and expletive were not significantly different. 
In comparisons between non-expletive and expletive speech, non-expletive was rated better in terms of \emph{discomfort} and \emph{likability}, and although trending was the same as in the initial study, the significant benefits of expletive over non-expletive remarks for \emph{social closeness} and \emph{humorousness} did not replicate.
As in the initial study, free-response input hinted at individual-level preferences for cursing. For example, one fan of the profanity said ``it was quite human-like and very funny, hearing the robots be emotional and curse.'' 
It seems clear from the results that a subset of people in the cultural context of the U.S. would value robot profanity, but non-expletive exclamations would be a safer singular design choice.

\vspace{-0.05in}
\section{In-Person \redwrap{Proof-of-Concept} Deployment}
\label{in_person}
\vspace{-0.05in}
Online studies can easily reach a large and diverse number of participants, but they have limited ecological validity. Accordingly, we also conducted an in-person proof-of-concept deployment to assess how the results of the online studies extend to situated interactions. We conducted this deployment in the wild via brief interactions wherein the Stretch robot handed out small gift bags to welcome students back to \redwrap{our U.S.-based university} campus for the Fall 2023 term. 

\vspace{-0.05in}
\subsection{Methods}
\vspace{-0.05in}
Methods for this between-subjects in-person \redwrap{deployment} were approved by the Oregon State University ethics board.


\label{methods 3}
\subsubsection{Study Design}
As with the online studies, we used the Stretch RE2 for our in-person deployment. 
We maintained the same three conditions for this study: no-speech, non-expletive, and expletive.
We wanted to create a fast and repeatable interaction in which the robot would make a mistake to pair with these conditions.
Accordingly, we constructed a wizarded interaction during which the robot would bump into the table in front of it.
We used tape marks on the ground to ensure consistent start and end poses for the robot, as well as position of the participant, in each interaction. The same trained operator controlled the robot during all interactions.
After bumping into the table, the robot would say nothing, say ``oops,'' or say ``d***it,'' depending on the pre-randomized condition; during early piloting, test users felt that this expletive matched the robot's error well, and it avoided the religious swear that stood out to online participants. 
All phrases used the same text-to-speech voice as in the online studies. The audio was played using Stretch's onboard speaker. 


\subsubsection{Measures}
To fit within the short time frame of the in-person deployment (but still capture similar metrics as in the online studies), we created a one-page on-paper survey that asked participants to rate the robot's \emph{friendliness}, \emph{service}, and \emph{funniness}, as well as the \emph{overall experience} with the robot (as proxies for the online survey measures). These questions were on 5-pt Likert-type scales for ease of fast completion.
In the on-paper survey, we also asked two yes-or-no manipulation check questions: whether the participant could hear the robot, and whether they thought the robot made a mistake. 
Lastly in the on-paper survey, we collected basic demographic information. 

The interactions with the robot were video-recorded, as were the brief (few-minute) closing interviews. We asked about general perception of the interaction with the robot, followed by question about the robot's relatability and any discomfort experiences during the interaction (as proxies for \emph{interpersonal closeness} and \emph{discomfort}, respectively).
As described later, we used video coding to support some understanding of \emph{humorousness} and \emph{anthropomorphism}.

\subsubsection{Procedure}
We conducted the study in an on-campus library space where passers-by could consent to participate by approaching the robot. 
One study team member sat behind the table with the robot in this space while another team member was visually obscured, remotely operating the robot. Consenting participants stood in front of the table while the robot approached to hand over a gift bag. The robot then bumped into the table, tilted the camera down to nonverbally acknowledge its mistake, and then either was silent or delivered one of the aforementioned phrases, depending on the condition. The participant took the gift bag, and then completed the on-paper survey and interview. 

\subsubsection{Participants}
Of the 58 willing participants who approached the table, six participants were excluded from analysis due to incomplete surveys or video malfunctions. This left us with 18 participants in the no-speech, 17 in the non-expletive, and 17 in the expletive condition (52 participants in total). Participants were 19 to 69 years old ($M=34.3$, $SD = 15.41$) and had some experience with robots ($\displaystyle M = 2.5$ out of 5, $SD = 0.98$). 67.3\% of participants had STEM training. No participants saw the interaction of the robot with someone else before completing the study. 
 
\subsubsection{Analysis}
\label{sec:analysis3}
We performed a Fisher's ANOVA (preferred for between-subject designs) on the scale data ($\alpha=0.05$). In the case of significant main effects, we used Tukey's HSD test. We report effect size for these results with $\displaystyle \eta^{2}$. 

We performed video coding on the study recordings, marking the occurrence of four key behaviors: describing the robot as relatable, describing discomfort during the interaction, laughing at the robot's reaction to the error, and using pronouns other than ``it'' when describing the robot.
We analyzed this binary data (occurring/not occurring for each given participant) with a non-parametric Kruskal-Wallis test ($\alpha=0.05$). When we found significance, we used a Dwass-Steel-Critchlow-Fligner pairwise comparison. We report effect sizes as $\displaystyle \epsilon^{2}$. 


\vspace{-0.05in}
\subsection{Results}
\label{Results 3}
\vspace{-0.05in}

This section details the results from the post-interaction survey and the video-coded results. 

\subsubsection{Survey Results}

The ANOVA tests for \emph{friendliness} ($p = 0.243$) and \emph{service} ($p = 0.523$) ratings yielded no significance. 
The ANOVA results for \emph{funniness} showed a significant main effect ($F(2,50) = 8.68, p<0.001$, $\displaystyle \eta^{2}$ = 0.262). Post hoc analysis revealed that the non-expletive ($M=4.3$, $SD = 0.82$) and expletive ($M=4.3$, $SD = 0.85$)  conditions were rated significantly higher than the no-speech condition ($M=3.2$, $SD = 0.94$).
ANOVA test results for \emph{overall experience} were not significant ($p = 0.898$). 
The manipulation check responses were always correct for the non-expletive and expletive conditions, but 22.3\% of users failed to identify the mistake for the no-speech condition. 

\subsubsection{Video Coding Results}
For \emph{relatability}, 16.7\% of participants said that the robot was relatable in the no-speech condition, compared to 64.7\% in non-expletive and 88.2\% in expletive. The Kruskal-Wallis test results showed significant main effects between the conditions ($\displaystyle \chi^{2} = 18.6, p<0.001$, $\displaystyle \epsilon^{2}$ = 0.365). Participants found the robot more relatable in the non-expletive and expletive conditions than in the no-speech condition. 
For \emph{discomfort}, 33.3\% of participants in the no-speech condition felt uncomfortable, compared to 41.2\% in non-expletive and 35.3\% in expletive. The Kruskal-Wallis test revealed no significant main effects between the conditions ($p = 0.885$). 
16.7\% of participants \emph{laughed} in the no-speech condition, compared to 70.5\% in non-expletive and 76.4\%  in expletive. The Kruskal-Wallis test showed significant main effects ($\displaystyle \chi^{2} = 11.3, p = 0.003$, $\displaystyle \epsilon^{2}$ = 0.222); the non-expletive and expletive conditions elicited more laughter than the no-speech condition. 
Lastly, for \emph{anthropomorphization}, 33.3\% of participants in the no-speech condition anthropomorphized the robot in some way, compared to 17.6\% in non-expletive and 52.9\% in expletive. The Kruskal-Wallis test results were not significant ($p = 0.194$). 

\vspace{-0.05in}
\subsection{Summary of Key Findings}
\vspace{-0.05in}
The information from the self-report questions and behaviors of participants show similar trends as the initial study. 
Having the robot respond to errors with speech was again beneficial to \emph{funniness}, \emph{relatability}, and \emph{laughter}, with the expletive condition tending to perform best in each of these areas. The manipulation check results also signal that using speech can help to highlight when an error has occurred, which may be useful in cases of human-robot collaboration. There were no significant differences between the non-expletive and expletive conditions. 
At the same time, among the participants, some were fans if the cursing (e.g., ``it was kind of amusing that I heard an expletive from a robot'') and others were not (e.g., ``I actually didn't like it and I am a swearer'').
Overall, results were similar to in the initial study, especially for relationship closeness and humor perceptions.

\vspace{-0.05in}
\section{Discussion and Conclusions}
\label{Discussion}
\vspace{-0.05in}




Between the three presented studies, there is strong evidence that verbal responses to failure are beneficial, and some evidence that cursing robots (especially with carefully-selected expletives) would gain a mostly positive perception in U.S. college campus contexts, although individual responses may vary. 
The findings on verbal responses to failure match well with past related literature (e.g.,~\cite{spatola2021ascribing, yam2021robots}). 
Our results also tended to signal that social norm violation in the form of cursing can elicit feelings of interpersonal closeness. 
In most comparisons, participants were unoffended by cursing robots and found them funny (rather than offensive), as in~\cite{yasuda2020perceived}, but this result was not universal, especially in the follow-up Prolific study.
At the same time, our outcomes seem to signal that robot profanity merits further exploration to clarify the ideal use contexts in which it offers additional value over non-expletive reactions, \redwrap{especially in cultural contexts in which swearing is most acceptable (e.g., North American, the United Kingdom, Australia, and New Zealand~\cite{hughes2015encyclopedia})}. \redwrap{We expect that this type of behavior might help to keep users interested or re-engage them, especially over long time horizons.} Beyond our work, there is also reason to challenge unconditional adoption of social norms in robotic systems; these norms can be outdated or paternalistic~\cite{coggins2023seven}.

At a high level, the \emph{design implications} of this work are that roboticists should: 1) continue leveraging verbal response to error as a helpful tactic for improving perception of robotic systems, and 2) consider uncouthness in the form of cursing as a shortcut for building rapport and as a helpful tactic for eliciting humor. The benefit of robots acknowledging their mistakes is already well known, but in an era of increasing emphasis on replication science, we consider our aligned results beneficial for the field of HRI. The more intriguing topic in this work is robot profanity. Between the first and third studies in this work, there were selected benefits, and almost no downsides, identified for the use of robot expletives. The second study, which involved a broad U.S. sample, resulted in slightly more precautionary results. We note that a lack of significance between non-expletive and expletive statements, as occurred in \redwrap{most} parts of our analysis, is an interesting result in itself; our instinct while entering this work was that we would encounter more downsides to cursing than we actually did. This outcome led us to the feeling that there is a latent ``continuum of mischief'' in HRI that merits further exploration, including (but not limited to) leveraging foul robot language to the benefit of user experience. \redwrap{We propose that exploring robot cursing in progressive spaces such as college campuses, as well as for personalized use deserves more attention in HRI. At the same time, certain precautions may need to be taken when deploying this type of technology; for example, using modern computer vision tools like DeepFace to identify minors who may be near the robot and avoid foul language when minors are detected.}

A \emph{strength} of this work is its relative uniqueness in terms of the use of cursing. 
The demonstration of replication for many of the results across the phases of the work also is a strength.
\emph{Limitations} of this work include the cultural context, \redwrap{few error types, short interactions, limited demographics, and small in-person deployment sample size.} 
We note the potential for rich follow-up work including \redwrap{experiments in more cultural contexts, additional error types (e.g., cognitive, moral, matched/unmatched to the situation) and contexts}, \redwrap{longer interactions}, more variety of swear words, additional robots, and \redwrap{more/more diverse participants}.
In our thinking, the first question to answer was how robot profanity is received (with a foundational series of studies), to decide whether more work on this topic is prudent. Based on the surprising finding that profanity was often equally acceptable to non-profane utterances, we propose that next exploration steps (including ideas listed above) are now merited.


In conclusion, this work introduced a series of investigations into \emph{cursing} as an expression of emotion, specifically when a robot fails. Our results generally support prior HRI findings on the benefits of mistake acknowledgment and norm-breaking, especially with careful expletive and use case selection.  Social robots that presently exist rely on being polite and careful not to offend; we show that this may not be necessary. We also make a case for robot personalization, as some people enjoy the uncouth robots and find them funny and enjoyable. Roboticists may benefit from using profanity as an engagement tool in personal and service robots.

\vspace{-0.05in}
\section*{Acknowledgments}
\vspace{-0.05in}
We thank Rafael Morales Mayoral for his support of the in-person proof-of-concept deployment. 


\vspace{-0.05in}
\bibliographystyle{IEEEtran}
\bibliography{myrefs}

\end{document}